\documentclass[12pt,journal,  draftclsnofoot,onecolumn]{IEEEtran}

\usepackage{url}
\usepackage{ifpdf}
\ifpdf
\usepackage[pdftex]{graphicx}
\else
\usepackage[dvips]{graphicx}
\fi

\DeclareGraphicsExtensions{.eps}
\usepackage{epstopdf}
\usepackage{tikz}
\usepackage{verbatim}
\usepackage{lipsum}
\usetikzlibrary{shapes,arrows}
\usepackage{caption}
\usepackage{subcaption}
\usepackage{graphics} 
\usepackage{epsfig} 
\usepackage{mathptmx} 
\usepackage{times} 
\usepackage{amsmath} 
\usepackage{amssymb}  
\usepackage{multicol}
\usepackage{multirow}
\usepackage{bm}

\DeclareMathOperator*{\argmin}{arg\,min}

\begin{document}
	
	\title{
		A Sparse Linear Model and Significance Test for Individual Consumption Prediction
	}
	
	
	\author{Pan Li,
		Baosen Zhang,
		Yang Weng,
		and~Ram~Rajagopal
		\thanks{Pan Li and Baosen Zhang are with the Department of Electrical Engineering, University of Washington, Seattle, WA, 98195, (e-mail: \{pli69, zhangbao\}@uw.edu).}
		\thanks{Yang Weng and Ram Rajagopal are with the Civil and Environmental Department, Stanford University, Stanford, CA, 94035, (e-mail: \{yangweng, ramr\}@stanford.edu).}
	}
	

	\maketitle


\begin{abstract}
	Accurate prediction of user consumption is a key part not only in understanding consumer flexibility and behavior patterns, but in the design of robust and efficient energy saving programs as well. Existing prediction methods usually have high relative errors that can be larger than 30\% and have difficulties accounting for heterogeneity between individual users. In this paper, we propose a method to improve prediction accuracy of individual users by adaptively exploring sparsity in historical data and leveraging predictive relationship between different users. Sparsity is captured by popular least absolute shrinkage and selection estimator, while user selection is formulated as an optimal hypothesis testing problem and solved via a covariance test. Using real world data from PG\&E, we provide extensive simulation validation of the proposed method against well-known techniques such as support vector machine, principle component analysis combined with linear regression, and random forest. The results demonstrate that our proposed methods are operationally efficient because of linear nature, and achieve optimal prediction performance.
\end{abstract}

\begin{IEEEkeywords}
	Load forecasting, least absolute shrinkage and selection, sparse autoregressive model, significance test
\end{IEEEkeywords}

\IEEEpeerreviewmaketitle

\section*{Nomenclature}
\addcontentsline{toc}{section}{Nomenclature}
\begin{IEEEdescription}[\IEEEsetlabelwidth{$\bm{\hat{\beta}_}{LASSO}$}]
	\item[$y_t$] Consumption at time $t$.
	\item[$\hat{y}_t$] Estimated consumption at time $t$.
	\item[$\beta_i$] Parameters in the regression model.
	\item[$\bm{\beta}$] Parameters as a vector.
	\item[$\epsilon_t$] Noise at time $t$.
	\item[$\mathbf{y}$] Consumption data as a vector.
	\item[$\mathbf{X}$] Regressor matrix in the regression model.
	\item[$\bm{\hat{\beta}_}{OLS}$] OLS estimate of $\bm{\beta}$.
	\item[$\bm{\hat{\beta}_}{LASSO}$] LASSO estimate of $\bm{\beta}$.
	\item[$\lambda$] Tuning parameter in LASSO.
	\item[$\mathbf{e}$] Fitted Residual after LASSO.
	\item[$\bm{\xi}$] Regressor matrix in the regression model for residual.
	\item [$\bm{\alpha}$] Parameters in the regression model for residual.
	\item [$ \bm{\psi}$] Noise at time $t$ in the regression model for residual.
	\item [$\bm{\hat{\alpha}}(\lambda_k)$] Estimated $\bm{\alpha}$ at tuning parameter $\lambda = \lambda_k$.
	\item[$\sigma^2$] Variance of the noise.
	\item[$\hat{\sigma}^2$] Estimated variance of the noise.
	\item [$F_1$] Test statistic.
	\item [$T_1$] Test statistic.
	\item [$F_{d_i,d_j}$] $F$ distribution with $d_i$ and $d_j$ as parameter.
	\item [$\mathcal{A}$] Active set.

\end{IEEEdescription}

\section{Introduction}\label{introduction}
\IEEEPARstart{E}{lectric} load forecasting is an important problem in the power engineering industry and have received extensive attention from both industry and academia over the last century. Many different forecasting techniques have been developed during this time. The authors in \cite{loadforecasting} present a comprehensive literature review on different methods related to load forecasting, from regression models to expert systems. Time series methods are further discussed in \cite{TaylorEtAl2007}. A thorough research on load and price forecasting is presented in \cite{Weron2007}. A common theme among many of the established methods is that they are used to forecast relative large loads, from substations serving megawatts to transmission networks serving more than gigawatts of power \cite{SevlianEtAl2014}. As the grid shifts to a more distributed system, the need of accurate forecasting for smaller sized loads is becoming increasing important.

Recent advances in technology such as smart meters, bi-directional communication capabilities and distributed energy resources have made individual households active participants in the power system. Many applications and programs based on these new technologies require estimating the future load of individual homes. For example, state estimation algorithms for distribution systems require pseudo-measurements \cite{BaranEtAl2009}, and these are provided by load forecasts. Another important class of application is demand response and dynamic pricing programs, where users' future demand are needed to design appropriate incentives \cite{MPC,ComPredict,DRmarket,Pan,Wang2016}.

In contrast to large aggregated loads, individual load forecasting is less developed. The current state of forecasting algorithms falls under three broad classes: simple averaging, statistical regression methods, and Artificial Intelligence (AI). They are listed in increasing order of prediction accuracy and decreasing in order of model simplicity. Simple averaging is intuitively pleasing since it is based on the mean of the previous similar days, but is often not very accurate. On the other end of the spectrum, AI methods can be extremely accurate, but it is difficult to associate the obtained parameter with the input data, i.e., past consumption. In this paper, we propose an algorithm that can achieve the performance of the state of the art AI methods, but retains the simplicity of linear nature of regression methods.

The algorithm we propose is based on the well known \emph{Least Absolute Shrinkage and Selection (LASSO)} algorithm in statistics and signal processing \cite{lasso1}.  It shrinks parameter estimates towards zero in order to avoid overfitting as well as pick up the most relevant regressors. Operationally, it minimizes the usual sum of squared errors, with a bound on the sum of the absolute values of the coefficients. LASSO is preferable in a setting where the dimension of the features are much higher than the size of the training set \cite{signconsistency2, lassomartin}. Furthermore, it can be easily extended to suit for more scenarios, i.e., group LASSO or adaptive LASSO \cite{Zou2006, Meier2008}. Due to its nice statistical properties and its efficiency, LASSO is applied to many different disciplines and has shown its superiority in sparsity recovery \cite{Xu2013, Bunea2014}. 

In particular, we formulate the load forecasting problem as learning parameters of a \emph{sparse} autoregressive model using LASSO. This sparse autoregressive model automatically selects the best recurrent pattern in historical data by shrinking irrelevant coefficients to zero. By selecting the correct features, the algorithm improves the order selection in autoregressive models, and as we will show using real load data from Pacific Gas and Electric Company (PG\&E), it also greatly improves the prediction accuracy of current regression models. For example, taking the Mean Absolute Percentage Error (MAPE) as a metric, autoregressive model with lag order one (AR(1)) has a MAPE of 33.9 \%, based on a pool of 150 users. The proposed method reduces the MAPE to 22.5 \%. Nonparametric methods such as using the consumption data at the same time during last week (LW method) are more intuitive but their performance are not robust subject to noise. They yield a relative error nearly up to 35\% based on a pool of 50 users in our test dataset. In addition, based on our dataset, linear regression model with the reduced feature space from Principle Component Analysis (PCA) performs rather poorly, with a relative error greater than 100 \%, which is even worse than simply predicting zero. Overall, the proposed model with LASSO achieves a much smaller error and maintains the intuitive nature of linear methods. 

In addition, we observe that given the set of coefficients, utilities and planners can easily identify and attribute the impact of different features on load consumption based on LASSO. For example, suppose for a given user, our method concludes that its load at time $t$ is mostly determined by the past load at times $t-1$, $t-3$ and $t-24$. This shows that this user has some short term behavior in the one hour range, medium term behavior in the 3 hour range, and a daily cycle partner that repeats every 24 hours. This information can be interpreted by utilities to better understand this user's consumption patterns and potentially identify appropriate demand programs. 

Apart from LASSO, data from other users can be leveraged for forecasting in the proposed framework. Intuitively, this means that knowing the past history of user $j$ improves forecasting of the user $i$.  For a given user, we use a sequential hypothesis testing procedure to find the best other user's historical data to include in the algorithm. We give a rigorous derivation of the hypothesis testing procedure and quantify the confidence of including other users. This allows us to show that the procedure is optimal, in the sense that given user $i$, it finds user $j$, whose historical data improves load forecasting for user $i$ the most among all users.

We derive rigorous theoretical justification for our methods as well as provide extensive simulation studies with respect to several well studied prediction methods. In particular, we compare against AutoRegression (AR), Exponential Smoothing (ES), Support Vector Machine (SVM), linear regression with PCA, Random Forest (RF) and Neural Network (NN) model. We also include two more nonparametric methods, i.e., using ten previous days' average consumption as prediction and using last week's (LW as mentioned earlier) consumption data as prediction. Using a user's own historical data, our proposed method and RF both reduce prediction error by 30\% compared to other predictors. Our proposed method is simpler in nature than RF since the latter is a generic machine learning technique that relies on a random ensemble of decision trees~\cite{Breiman2001,Ho1995}. Therefore, our method is useful to system operators in policy decisions without sacrificing prediction accuracy. By adding the historical data of another user, we can further improve the prediction accuracy.

\subsection{Contribution}\label{Cont}
The contribution of this paper is two fold. 
\begin{itemize}
	\item First, we apply LASSO in consumption prediction. LASSO has been studied in literature to jointly predicts price, load and renewables together \cite{priceforecasting}. However, there is little discussion on how LASSO can be applied to consumption forecasting without using other side information. Our paper has a thorough discussion on how LASSO is used in obtaining a sparse autoregressive model and compares its performances with several other popular prediction methods in literature. The simulation results show that LASSO achieves competitive prediction performance compared to nonlinear machine learning algorithms such as RF or NN. In addition, LASSO is computationally much faster and easier to be understood by human operators.
	
	\item Second, we propose a significance test that can leverage other users' consumption data for prediction. This testing procedure differs significantly from standard clustering algorithms since it looks for the most ``predictive'' user, not necessarily the most similar user. 
\end{itemize}

%

The rest of the paper is organized as follows. Section \ref{LR} analyzes related work in short term load forecasting. Section \ref{autoregressive} presents the autoregressive model for time series analysis. Section \ref{Sparisty} introduces LASSO type linear regression model. Section \ref{multi} proceeds with the significance test to pair users in order to improve prediction performance. It describes the significance test for LASSO, i.e., covariance test, to select the most significant user to form the pair with the current user. Section \ref{simulation} introduces the evaluation criterion for prediction and details the simulation of the proposed methods compared to several other popular prediction methods. And finally Section \ref{conclusion} concludes the paper and draws avenues for future work.


\subsection{Literature review}\label{LR}
There exists an extensive literature on short term load forecasting \cite{tao1,loadforecasting}. In summary, research on energy consumption prediction can be divided into three groups \cite{loadforecasting}, including simple averaging models, statistical models and AI models.


The simplest approach is to employ moving average \cite{ISO}. Such models make predictions on mean of consumption data from previous similar days \cite{bsload}. AI type methods (e.g. NN or RF) yield high accuracy at the cost of complexity of the system, which may lead to overfitting \cite{AI}. Other drawbacks include difficult parametrization and non-obvious selection of variables. Statistical methods sit in between the previous methods in terms of complexity and accuracy, and include regression models, probabilistic approach applied to regression models, and time series analysis such as autoregressive models.

In statistical methods, regression models combine several features to form normally a linear function. In \cite{forecast}, the authors build a regression tree model with weather data to predict consumption. SVM is used in \cite{SVM}. Gaussian process framework for prediction mitigating the uncertainty problem is proposed recently in \cite{YangGaussian}.

Advanced AI models have also been used to facilitate load forecasting. Authors in \cite{Quan2014} propose a NN model to handle uncertainties associated with forecasts. Similarly, authors in \cite{Sarwat2016} study the interruption prediction with weather conditions by simulating a NN. Besides these work, the authors in \cite{Fiot2015} explore kernel-based learning techniques to forecast both commercial and industrial load. In addition, feature selection techniques are applied in \cite{Abed2016} to predict electricity load and prices.

Apart from these advanced machine learning tools, another type model involves time series based methods \cite{ref1}. An overview can be found in \cite{Signal}. Models such as ES \cite{ES2012} and the Autoregressive Integrated Moving Average (ARIMA) model \cite{tsforecast,ref2} are all time series based. ARIMA is a widely used time series based prediction methods and have been adopted in \cite{Contretas2003,Amini2016}. In these papers the authors apply ARIMA to predict either the electricity price or the electric vehicle demand. Apart from these work, authors in \cite{priceforecasting} propose a vector autoregressive model to include renewables, prices and loads together with sparsity recovery. In this work, the authors have also explored LASSO to obtain a sparse linear model. In addition, to extend from linearity to nonlinearity,  \cite{clusteringnew} addresses a mixed model combining ARIMA model to deal with the linear part and neural network with the nonlinear one. In our work, we recover the sparsity for univariate time series and multivariate time series under the framework of autoregressive models.

\section{Autoregressive model}\label{autoregressive}
Autoregressive models are widely used for prediction and inference of time series data. Here we adopt a linear autoregressive model of the hourly consumption of a single user, where future demands $y_t$ depend linearly on historical data $y_{t-i}$ plus random noise:
\begin{equation}\label{autoregressive model}
y_{t} = \beta_{0}+ \sum_{i=1}^{I}{\beta_{i}y_{t-i}} + \epsilon_{t}.
\end{equation}

In this model, $y_t$ denotes the demand of the user at time $t$, $\beta_i$ is the coefficient for order (lag order) $i$ in the autoregressive model (it represents the weight of each historical demand data $y_{t-1}$ in predicting future demand $y_t$), and $\epsilon_t$ is an additive random Gaussian noise. The time index $t$ is measured in hours and the noise is identically and independently distributed at different hours. Note that in this paper we denote time series data by notation $\{\bullet_t\}$, where subscript $t$ refers to the time slot in this time series data. In addition, $I$ is the number of orders that we include in the model.  An autoregressive model with maximum order $I$ is denoted by AR($I$).



To use the model in \eqref{autoregressive model} for prediction, the standard approach is to use Ordinary Least Squares (OLS) to estimate the coefficients $\beta_i,\; i \in I$. By convention, we write  $[y_{t-1}\quad y_{t-2}\quad \cdots \quad y_{t-I}]^{\text{T}}$ as a vector denoted by $\bf{x}_t$. Using this notation, the model in \eqref{autoregressive model} is written in a compact matrix form:
\begin{equation}\label{linear regression}
\mathbf{y} = \mathbf{X}\bm{\beta} + \bm{\epsilon},
\end{equation}
where $\bf{y}$ = $[y_t\quad y_{t+1}\quad \cdots]^{\text{T}}$, $\mathbf{X}$ is a matrix where $t^{\text{th}}$ row is $[1 \quad \bf{x}_t^\text{T}]$,  $\bm{\beta}$ = $[\beta_0 \quad \beta_1\quad \cdots]^{\text{T}}$, and $\bm{\epsilon}$ = $[\epsilon_t \quad \epsilon_{t+1}\quad  \cdots]^{\text{T}}$. Vectors $\bf{y}$, $\bm{\beta}$ and $\bm{\epsilon}$ have dimension $T$. Matrix $\mathbf{X}$ has $P$ columns, which we refer to as the dimension of $\mathbf{X}$.

Applying standard OLS to \eqref{linear regression}, the estimate of $\bm{\beta}$ is given by:
\begin{equation}\label{OLS}
\bm{\hat{\beta}_}{OLS}= \argmin_{\bm{\beta}}\left\Vert\left(\bf{y-X}\bm{\beta}\right)\right\Vert_2^2.
\end{equation}


Under some standard assumptions, the OLS estimator $\bm{\hat{\beta}_}{OLS}$ is a consistent estimator for the true $\bm{\beta}$, meaning that the expected difference between the estimator and the true value approaches zero when sample size becomes large~\cite{tsOLS}. This means that the bias goes to zero as the sample size becomes big. The other asymptotic analysis on the OLS also applies in this case, such as asymptotic distribution for gaussianality of estimators and significance tests; however, in autoregressive models the estimators are typically not unbiased.

To learn and evaluate the estimator $\bm{\hat{\beta}_}{OLS}$ from OLS, we separate the dataset into a training set and a test set. The estimator $\bm{\hat{\beta}_}{OLS}$ is learned from the data from the training set and the estimation error is evaluated on the test set. Note that estimators may exhibit extremely good fits on the training set but poor estimation performance on the test set, as according to~\cite{hastie}.

\section{Sparsity in Autoregressive Models}\label{Sparisty}


Since the objective of OLS estimators is to minimize the sum of squared errors in the training set, OLS achieves optimal \emph{in-sample} performance. This means that adding more regressors into (\ref{linear regression}) can always decrease the sum of squared error and better fit the data within the training set. However, when we include too many irrelevant regressors, i.e. when we include too many lag orders from the historical data in (\ref{linear regression}) , we are misled by the reduced in-sample bias. We will then ignore the high variance introduced by the estimator which leads to model overfitting. 

\begin{figure}[!t]
	\centering
	\includegraphics[width=0.8\columnwidth]{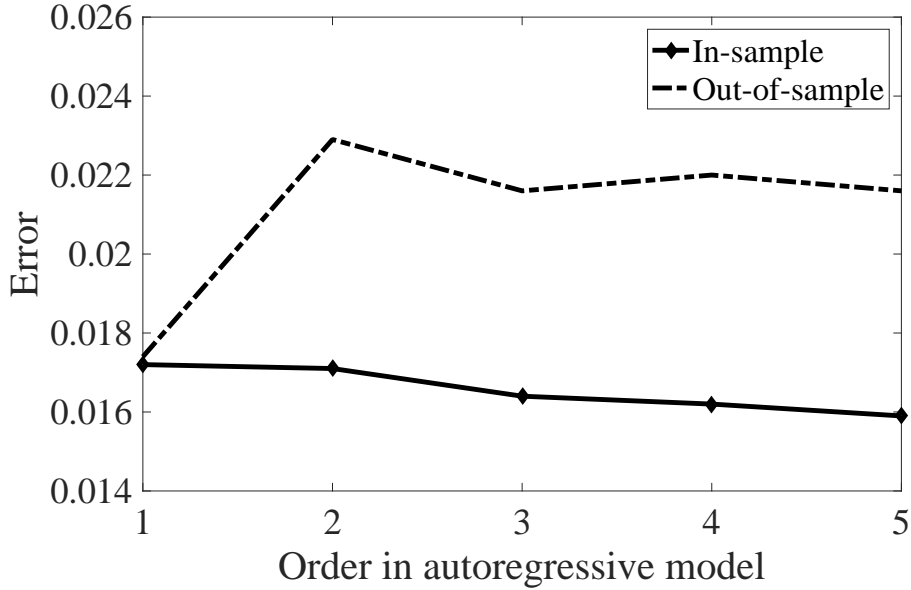}
	\caption{Error comparison between different autoregressive models.}
	\label{error}
\end{figure}

We use the PG\&E dataset as an illustrative example. In this particular dataset, hourly consumption data for single households is recorded. If we use an AR(5) model for a particular household, it will result in an average in-sample squared error of 0.0159, with an average out-of-sample error of 0.0216, whereas AR(1) model has an average in-sample squared error of 0.0172, together with an average out-of-sample error of 0.0174. Thus AR(1) gives better out-of-sample fitting results. If the potential lag orders are up to 10 days, i.e., 240, then an AR(240) model would produce large out-of-sample errors. Overall, we need to select the lag orders carefully to avoid model overfitting. The tradeoff between out-of-sample error and the order of the autoregressive model is shown in Fig. \ref{error}. 

Determining the correct lag orders is not trivial because it is a combinatorial problem, which is NP-hard. To this purpose, we use LASSO, which is a convex relaxation of such combinatorial problems, to select relevant lag orders. The intuition for using LASSO is to get a sparse autoregressive models with high orders. 



More formally, LASSO is a shrinkage and selection method for linear regression described in (\ref{linear regression})~\cite{lasso1}. It shrinks parameter estimates towards zero in order to avoid overfitting as well as picks up the most relevant regressors. Operationally, it minimizes the usual sum of squared errors, with a bound on the sum of the absolute values of the coefficients:
\begin{equation}\label{LASSO}
\bm{\hat{\beta}_}{LASSO} = \argmin_{\bm{\beta}}\frac{1}{2}\left\Vert\left(\bf{y-x}\bm{\beta}\right)\right\Vert_2^2 + \lambda \left\Vert\bm{\beta}\right\Vert_1,
\end{equation}
where $\lambda$ is a tuning parameter to control the level of sparsity in the solution. In essence, it controls the number of past consumptions included in the prediction of future consumptions (i.e. sparsity). The bigger $\lambda$ is, the more sparse the solution $\bm{\hat{\beta}_}{LASSO}$ is, which means that less number of data points from the past is used to predict the future. This parameter balances the training and testing performances of the model. When $\lambda = 0$, the solution is the same as in (\ref{OLS}) as a traditional autoregressive model. For practical purposes, we are using $k$-fold cross validation (CV) to determine the value of $\lambda$ in our simulations, where $k$ is either 5 or 10~\cite{hastie}.

LASSO has gained wide spread popularity in signal processing and statistical learning, see ~\cite{lassosig1,lassosig2,lassosig3}. LASSO has also been applied to forecast electricity price~\cite{priceforecasting, pricelasso}, but its application to load forecasting is still a new topic. In~\cite{loadlasso}, LASSO has been applied to forecast short and middle term load. In our paper, we adopt LASSO to predict hour-ahead consumption and provide a comparison between LASSO and some other well-used prediction methods in literature. As discussed later in Section \ref{simulation}, LASSO is the most preferable model when considering model simplicity and prediction performances. Despite the fact that RF achieves the best prediction performance with an average relative error less than 20\%, it is a highly non linear model and requires to tune many hyper-parameters. This is computationally expensive and not linear in nature. On the other hand, LASSO achieves similar performance as RF and outperforms all other linear models considered in this paper, i.e., SVM with a linear kernel, linear regression with PCA, AR(1) and simple averaging method. It reduces the relative error by 30\% compared to these models. Thus LASSO is considered the best in terms of simplicity and prediction accuracy based on our dataset.

Another advantage of using LASSO to recover sparsity is that it has some nice properties as to sign consistency~\cite{signconsistency2,lassomartin}. This means that $\hat{\bm{\beta}}$ has the same support as $\bm{\beta}$ and the sign of each element in this support is recovered correctly. Therefore LASSO recovers the exact sparsity of the underlying model. In our simulation, LASSO selects both the most recent lag orders and lag orders with intervals of roughly 24 hours, which performs as a combination of simple averaging and AR($\cdot$). Furthermore, LASSO also gives more intuitive results with respect to selected orders. In our simulation, for one electricity user as an example, LASSO selects lag orders as 1, 2, 5, 6, 16, 22, 23, 24, 48, 143, 144, 160, 191, 216, 238, 240. From these orders we can observe a clear behavior pattern of an interval of 24 hours. Some are multiples of 24 and some are not but close to multiples of 24. We thus observe that not every lag order that LASSO picks is a multiple of 24, otherwise we would directly employ simple averaging rather than LASSO, so LASSO is more adaptive and flexible than simple averaging or AR($\cdot$). This implies that user behavior at current hour depends on similar hours happened in previous days. Unlike simple averaging which fix the lag orders at 24, 48, 72, etc., LASSO will automatically select these orders for each individual based on their respective historical data, instead of directly imposing fixed orders.

As can be seen from Section \ref{simulation}, LASSO applied to autoregressive model achieves the best prediction performance among all the linear models considered in this paper, with a relative error as small as 22.5 \%. It improves the prediction performance by 33.6 \% as compared to an AR(1) model.

\section{User Pairing by Significance Test}\label{multi}
So far we have considered using historical data of an individual user for its own prediction. One way to leverage the fact that we have many users' data is to improve the univariate autoregressive model by including other user's historical data into the model. One popular way to perform this is to conduct Vector Autoregression (VAR), which extends the univariate autoregressive model to joint prediction for a vector of time series data. 

To perform a complete VAR, we need to include all potentially relevant users into the autoregressive model, which will reduce the bias but increase the variance for estimators. This causes the same overfitting problem as occurred in univariate autoregressive model, when AR(3) yields a worse prediction on the test set but a better fit on the training set compared to AR(1). One possible way to overcome this problem is to first cluster similar users together and then perform VAR for each cluster. However, consider a scenario where two time series have the exact same values for each time slot. Then these two time series are clustered together since they are identical. In this case, clustering them together and performance multivariate autoregressive model does not help to enhance prediction because knowing the history of one time series would not help predict the future values of another time series. This problem distinguishes similarity from prediction performance, which is the focus of this section.

In this paper we focus on selecting the most relevant user to enhance prediction performance after doing univariate LASSO selection. To this end, we adopt LASSO significance test to select this most relevant user. In LASSO significance test, the inclusion of a particular user is based on how well this particular user's history data explains the fitted residual after performing LASSO to one user's univariate autoregressive model. This is a hypothesis test of an exponential random variable~\cite{lassotest}. We will discuss more details of a LASSO solution to a regression model and the implementation of the LASSO significance test in the rest of this section. Overall from the simulation results presented in section \ref{simulation}, using significance test on top of univariate LASSO-type regression model improves the relative prediction error from 22.5\% to 20.9\%, which is almost as good as the prediction results from RF. 




\subsection{Linear Regression Model for the Fitted Residual and LASSO Path}

In LASSO significance test, we want to test if the fitted residual from LASSO solution in (\ref{LASSO}) is indeed noise or if it can be better explained by other user's historical data. Intuitively, We need to test inclusion of each user's historical data. We therefore generate LASSO autoregressive coefficients for univariate time series in (\ref{autoregressive model}) and compute the residual $\{e_t\}$ (true value minus fitted value) for each individual user. Then we use $e_t$ as the response variable in a new linear regression model. To fit this new linear regression model, we include all users' historical data at lag order one except for this user as the regressors. In this way we will have a high dimensional regressor matrix. Its dimension is the number of users we want to potentially include, which is the total number of users minus one. Since we only want to include the most significant user as the regressor, LASSO is therefore performed and furthermore tested on this new linear regression problem. In the following, we first illustrate more details on formulating the LASSO regression model to the fitted residual from univariate model. Then we provide more explanations on the LASSO path with a varying $\lambda$. In the next subsection we introduce the covariance statistic and its asymptotic distribution for hypothesis tests along the LASSO path.

Mathematically, the new linear regression problem for the residual is formulated as follows:
\begin{equation} \label{residual}
\mathbf{e} = \bm{\xi}\bm{\alpha} + \bm{\psi},
\end{equation}
where $\mathbf{e} = [e_{t,s}\quad e_{t+1,s}\quad \cdots]^{\text{T}} = [y_{t,s} - \mathbf{X}_{t,s}^{\text{T}}\bm{\hat{\beta}}_{LASSO}\quad y_{t+1,s} - \mathbf{X}_{t+1,s}^{\text{T}}\bm{\hat{\beta}}_{LASSO}\quad \cdots]^{\text{T}}$ is the residual for the current user $s$,  $\mathbf{X}_{t,s}$ = $[y_{t-1,s}\quad y_{t-2,s}\quad \cdots \quad y_{t-I,s}]^{\text{T}}$ and $y_{t,s}$ denotes the consumption data for user $s$ at time $t$. The parameter $\bm{\alpha}$ is a vector of weights that denotes how helpful each other user's last demand data is towards the prediction of future demand of user $s$. Vector $\mathbf{e}$ has a length of $N$. Moreover, matrix $\bm{\xi}$ is the regressor matrix made by lag order one historical data from all the other users except for the current user $s$. The $t^{\text{th}}$ row of $\bm{\xi}$ is $[y_{t-1,1} \quad y_{t-1,2} \cdots y_{t-1,s-1} \quad y_{t-1,s+1} \cdots]$. So $\bm{\xi}$ has a column length of $P$, which is the number of users to be included, and a row length of $N$, which is the same as that of $\mathbf{e}$. One user is represented by one column of $\bm{\xi}$ and is regarded as one regressor variable. In addition, $\bm{\psi}$ is the white noise vector with variance $\sigma^2$. The parameter $\bm{\alpha} = [\alpha_0\quad \alpha_1\quad \cdots \quad \alpha_P]^{\text{T}}$ is the decision variable for the regression problem.

We again apply LASSO for (\ref{residual}), to avoid overfitting by including too many irrelevant users. Building on the discussion in Section \ref{Sparisty}, we define the LASSO path as the revolution of the estimator $\bm{\hat{\alpha}}$ in terms of a sequence of $\lambda_k$'s.
The LASSO path $\hat{\alpha}(\lambda_k)$ is given by:
\begin{equation} \label{LASSOpath}
\bm{\hat{\alpha}}(\lambda_k) = \argmin_{\bm{\alpha}}\frac{1}{2}\left\Vert\left(\mathbf{e}-\bm{\xi}\bm{\alpha}\right)\right\Vert_2^2 + \lambda_k \left\Vert\bm{\alpha}\right\Vert_1,
\end{equation}
where $\lambda_k$ is called the knot along the LASSO path.



For different values of $\lambda_k$, we obtain different solutions and sparsity at different levels. The active set at one particular value of $\lambda_k$ is the set of all non zero coefficients estimated at that value, i.e., $\mathcal{A}\ = \{\hat{\alpha}_p\neq0, \lambda = \lambda_k, p = 0, 1, \cdots, P \}$. The path $\hat{\bm{\alpha}}(\lambda_k)$ is continuous and piecewise linear with knots at these values $\lambda_1\geqslant\lambda_2\geqslant \cdots \geqslant0$~\cite{lasso2012}. With the formulation in (\ref{residual}), the goal is to test if an inclusion of one user's historical data is helpful for prediction. Mathematically speaking, we want to test if the variables that sequentially enter the active set are statistically significant.


\subsection{Covariance Test}

Significance test applied to the LASSO path is fundamentally different from the standard hypothesis tests in linear regression. In standard testings with linear regression model, we fix a regressor variable and test the significance of its inclusion into the model. This can be done by $t$-test. However, the scenario is not the same in the case where we want to test the significance of variables along the LASSO path since the variables entering the active set are not known in advance. This means that we do not know which regressors to fix in order to conduct $t$-test. This problem is addressed by the authors in~\cite{lassotest}. Instead of standard $t$-test, they propose a significance test for regressors progressively selected by the LASSO path, using the so called covariance statistic. They have shown that the covariance statistic asymptotically follows exponential distribution and can be used to test the significance of the entry of variables into the active set.

Using the covariance test defined in~\cite{lassotest}, we test whether it is significant to include one variable (one other user) into the active set against including no variables into the active set. Specifically, we want to analyze whether the residual, i.e., $\mathbf{e}$ is white noise, or it can be explained significantly by the lag one order historical data from other users.

The null hypothesis is that the true active set is an empty set, i.e., no other users can help better predict the current user:
\begin{equation} \label{globalnull1}
H_0: \alpha_1 = \alpha_2 = \cdots = \alpha_P = 0.
\end{equation}

Another way to interpret this is:
\begin{equation} \label{globalnull2}
H_0: \mathbf{e}\overset{H_0}{\sim}N(0, \sigma^2I).
\end{equation}

We focus on testing the significance of the first variable entering the active set and the two main reason for doing so are as follows. First, for the following variables entering the active set, the exponential distribution turns out to yield a slightly higher value than the true value of the test statistic, so the decision tends to be conservative~\cite{lassotest}. Second, If one would like to include more significant regressors selected by LASSO, the Tailstop criterion~\cite{sequentialsig} for ordered selection can be used under the assumption that sequential $p$-values are independent. However, we restrain from looking at $p$-values at greater indices because of the non-orthogonality of $\bm{\xi}$ in our case (which leads to misleading interpretation of $p$-values), as pointed out in~\cite{sequentialsig} and~\cite{discussionLASSOtest}. To illustrate this, we use Tailstop criteria in our simulation and it turns out that the first eighteen regressors are selected to limit the type I error of the global null in (\ref{globalnull1}) under 0.05. It is obviously an overfitting result and Tailstop criterion cannot be trusted due to the correlation between each row in the regressor matrix $\bm{\xi}$. From literature, interpreting $p$-values from sequential hypothesis and order selection under generic conditions is still an open question and is thus beyond the discussion of this paper.

To test whether the first variable entering the active set is significant, we first set $\lambda = \infty$ and gradually reduce $\lambda$ until one regressor variable has a non-zero $\alpha$. Denote the value of this $\lambda$ as $\lambda_1$. Also denote $\lambda_2$ as the value of $\lambda$ when the second regressor variable enters the active set. For simplicity of representation, we scale the columns of the $\bm{\xi}$ so that each column has unit norm.

Following this scaling strategy, the authors in~\cite{lassotest} define the covariance statistic for testing the entry of the first variable and it is written as:
\begin{equation}\label{T}
T_1 = \frac{\lambda_1(\lambda_1-\lambda_2)}{\sigma^2}.
\end{equation}

From Theorem 2 in~\cite{lassotest}, the authors have proved that $\Pr \{T_i >t\} \rightarrow e^{-t}$ as $P \rightarrow \infty$ for all $t\geq 0$. Thus we have that the asymptotic distribution of $T_1$ obeys exponential distribution with parameter 1: $T_1 \rightarrow \text{Exp}(1)$.

In most cases, the value of $\sigma^2$ is not known a priori.
We can estimate $\sigma^2$ by the residual sum of squared error via:
\begin{equation}\label{errorestimation}
\hat{\sigma}^2 =\frac{ \left\Vert\left(\mathbf{e}-\bm{\xi}\hat{\bm{\alpha}}_{OLS}\right)\right\Vert_2^2}{N-P},
\end{equation}
where $\bm{\hat{\alpha}}_{OLS}$ is obtained through:

\begin{equation}\label{OLScov}
\bm{\hat{\alpha}}_{OLS} = \argmin_{\bm{\alpha}}\frac{1}{2}\left\Vert\left(\mathbf{e}-\bm{\xi}\bm{\alpha}\right)\right\Vert_2^2.
\end{equation}

Plugging (\ref{OLScov}) into (\ref{T}), we have a new statistic $F_1$, which is asymptotically following $F$ distribution (ratio of two independent $\chi^2$ distribution) under the null~\cite{lassotest}:
\begin{equation}\label{F in LASSO}
F_1 = T_{1}\frac{{\sigma}^2}{\hat{\sigma}^2} = \frac{\lambda_1(\lambda_1-\lambda_2)}{\hat{\sigma}^2} \to F_{2, N-P}.
\end{equation}

In conclusion, to test whether another user's historical data can explain the residual obtained by the univariate LASSO introduced in Section \ref{Sparisty}, we compute the value of $\lambda_1$ and $\lambda_2$, along with the full linear regression in (\ref{OLScov}) which gives us $\hat{\sigma}^2$. Plugging these into (\ref{F in LASSO}), we include the regressor variable (which represent one particular user) entering the active set at $\lambda_1$ if $F_1$ is greater than some threshold and reject the null hypothesis in (\ref{globalnull2}). Simulation results show that performing this significance test on top of univariate LASSO type regression model has improved the prediction by reducing relative test error by 38.3\%, compared to AR(1) model.


\section{Simulation Results}\label{simulation}

We use the data from PG\&E. It contains hourly smart meter readings for residential users during a period of one year from 2010 to 2011. Temperature data is retrieved from an online database~\cite{weblink} for the same period. Some preliminary observations suggest that temperature is not a significant regressor to predict consumption based on our data set, thus it is excluded from the regressors for the following simulations. In addition, the consumption data has been filtered to exclude daily periodic trends and is separated into weekday data and weekend data. The simulation results are obtained on weekday data.

\subsection{Prediction Evaluation Criteria}\label{EvaluationCr}
Before proceeding with the prediction methods, we first introduce the evaluation criteria for prediction used in this paper. A naive way to evaluate prediction methods is to compare the Mean Squared Error (MSE) or Mean Absolute Error (MAE) within the testing set:
\begin{equation}\label{MSE}
\mathrm{MSE} = \frac{1}{n}\sum_{t=1}^{n}(y_t-\hat{y}_t)^2
\end{equation}

\begin{equation}\label{MAE}
\mathrm{MAE} = \frac{1}{n}\sum_{t=1}^{n}|y_t-\hat{y}_t|
\end{equation}
where $y_t$ is the actual value, $\hat{y}_t$ is the predicted value, and $n$ is the number of fitted values.

However,  MSE and MAE are not scale invariant. This will be misleading if we want to compare the performance for two datasets with significantly different scales and means. One way to solve this problem is to normalize the error or to compare the relative error, i.e., the prediction error with respect to the data scale. Here we use the MAPE mentioned earlier in Section I, to capture the relative error:
\begin{equation}\label{MAPE}
\mathrm{MAPE} = \frac{1}{n}\sum_{t=1}^{n}|\frac{y_t-\hat{y}_t}{y_t}|.
\end{equation}

Normalized Root-Mean-Square Deviation (NRMSD) is also a good metric to compare prediction performances of data with different scales :
\begin{equation}\label{NRMSD}
\mathrm{NRMSD} = \frac{\sqrt{\frac{1}{n}\sum_{t=1}^{n}(y_t-\hat{y}_t)^2}}{\bar{y}},
\end{equation}
where $\bar{y}$ is the mean of $y$ in the training set.

In case of outliers, we adopt mean curtailing of 0.01 tail and head, or simply use the median to replace the mean value.

In addition, to evaluate the model between good fit and complexity (in terms of the number of parameters), Akaike Information Criterion (AIC) and Bayesian Information Criterion (BIC) can be used \cite{Kia17}. 

\subsection{Comparison with Different Prediction Methods}\label{section_compare}

We compare our proposed method with other popular methods: 1): Simple averaging, which averages the consumption at the same hour of the day for ten previous days; 2): LW, which uses the consumption of the same hour of the day during last week;  3): AR(1); 4): ES with additive error; 5): linear regression with PCA, 6): SVM with and without PCA 7): RF and 8): NN with a single hidden layer. Input features are ten previous days' consumption data which has a dimension of 240, except that for NN the input features are set to be previous day's consumption data, which has a dimension of 24. While the first and the second method are very intuitive methods for prediction, AR(1) and ES are popular prediction methods in analyzing time series data. The remaining methods are popular machine learning algorithms and artificial intelligence techniques in prediction. Particularly RF and NN are the only non linear methods considered in this paper. They serve to provide reference results as compared to other linear methods. Since the dataset that we use only contains time series load consumption data (plus coarse zip code data), we restrain from using more complexed models which tend to overfit. In all, our simulations using real data show that our proposed method performs similar or better to these sophisticated algorithm while remaining a simple linear structure.


We make the following clarifications for the machine learning techniques in 5)-8). First, for the extracted features obtained from PCA, we threshold the number of components with a tolerance $\zeta$ on how much the chosen components explain the covariance matrix of data $\bm{\xi}$. Then we include those components as new features to the linear regression model. If we set up a tolerance $\zeta = 0.75$ and omit components if their standard deviations are smaller than $\zeta$ times the standard deviation of the first component, we obtain an average two principle components for each user based on our dataset. Second, a linear kernel is trained as we are comparing SVM with several linear regression models. Third, for RF, we grow 500 trees in total. At each split of the tree node we re-sample a third of the original features. Note that RF is a nonlinear model in terms of input features whereas the other models are linear in input features. For a more in-depth introduction to RF, please refer to~\cite{RF}. Last, due to convergence and computational issues, we consider one hidden layer with five neurons in the NN model. 

\begin{figure}[!t]
	\centering
	\includegraphics[width=0.9\columnwidth]{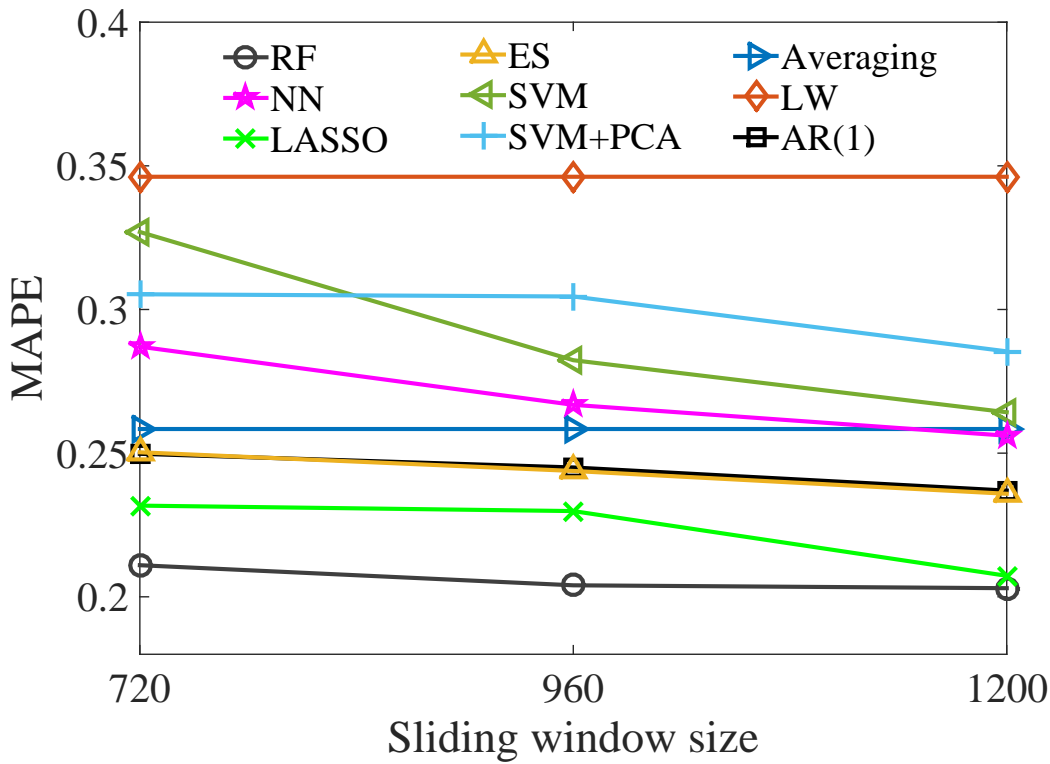}
	\caption{MAPE for different prediction methods with various window size of the training set. RF yields the best prediction performance among all aforementioned methods and LASSO yields the best performances among linear models.}
	\label{compare}
\end{figure}

\begin{figure*}
	\centering
	\begin{minipage}[b]{0.4\linewidth}
		\centering
		\includegraphics[
		width=1\textwidth]{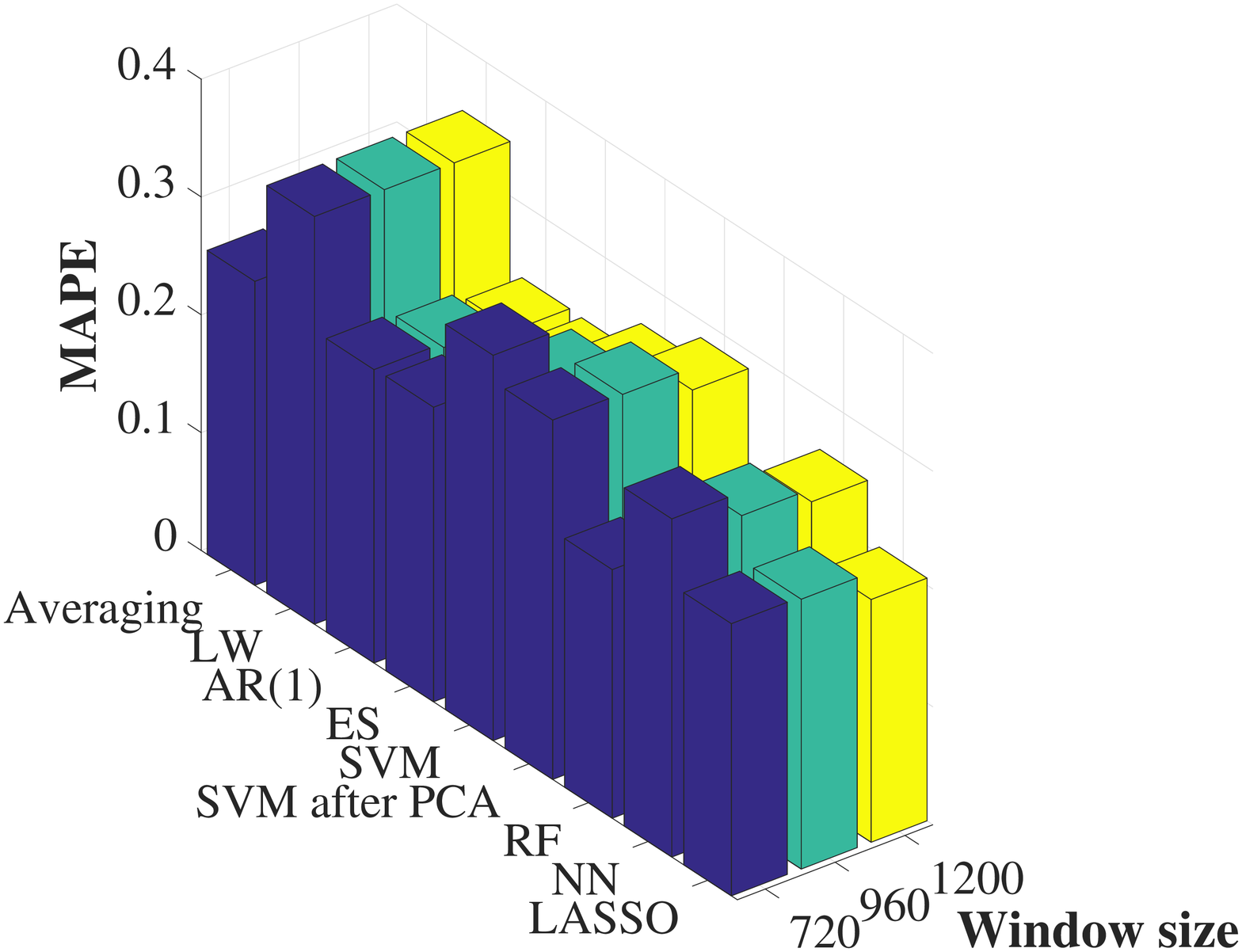}
		\subcaption{Average MAPE.}
		\vspace{4ex}
	\end{minipage}
	\begin{minipage}[b]{0.4\linewidth}
		\centering
		\includegraphics[
		width=1\textwidth]{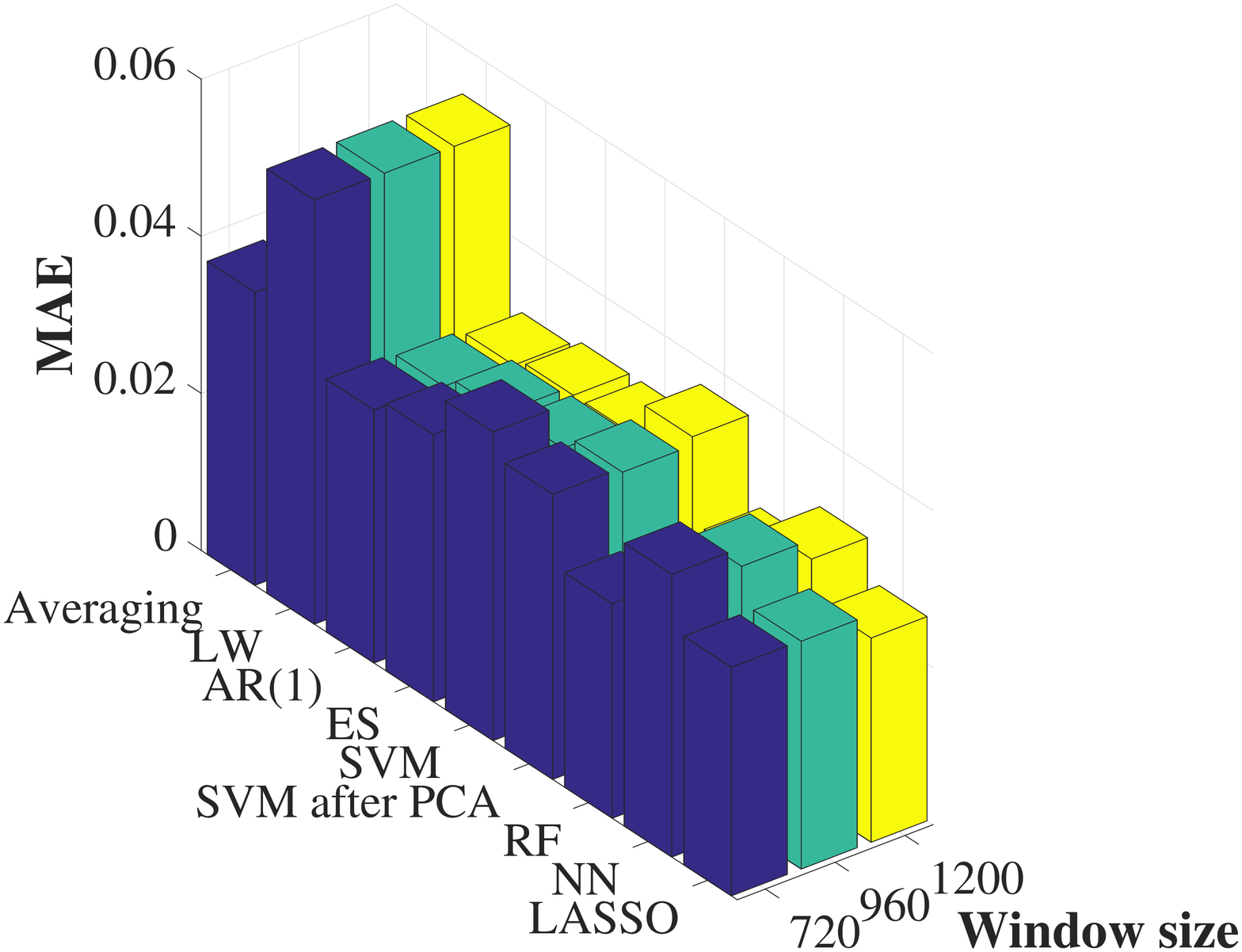}
		\subcaption{Average MAE.}
		\vspace{4ex}
	\end{minipage}
	\begin{minipage}[b]{0.4\linewidth}
		\centering
		\includegraphics[
		width=1\textwidth]{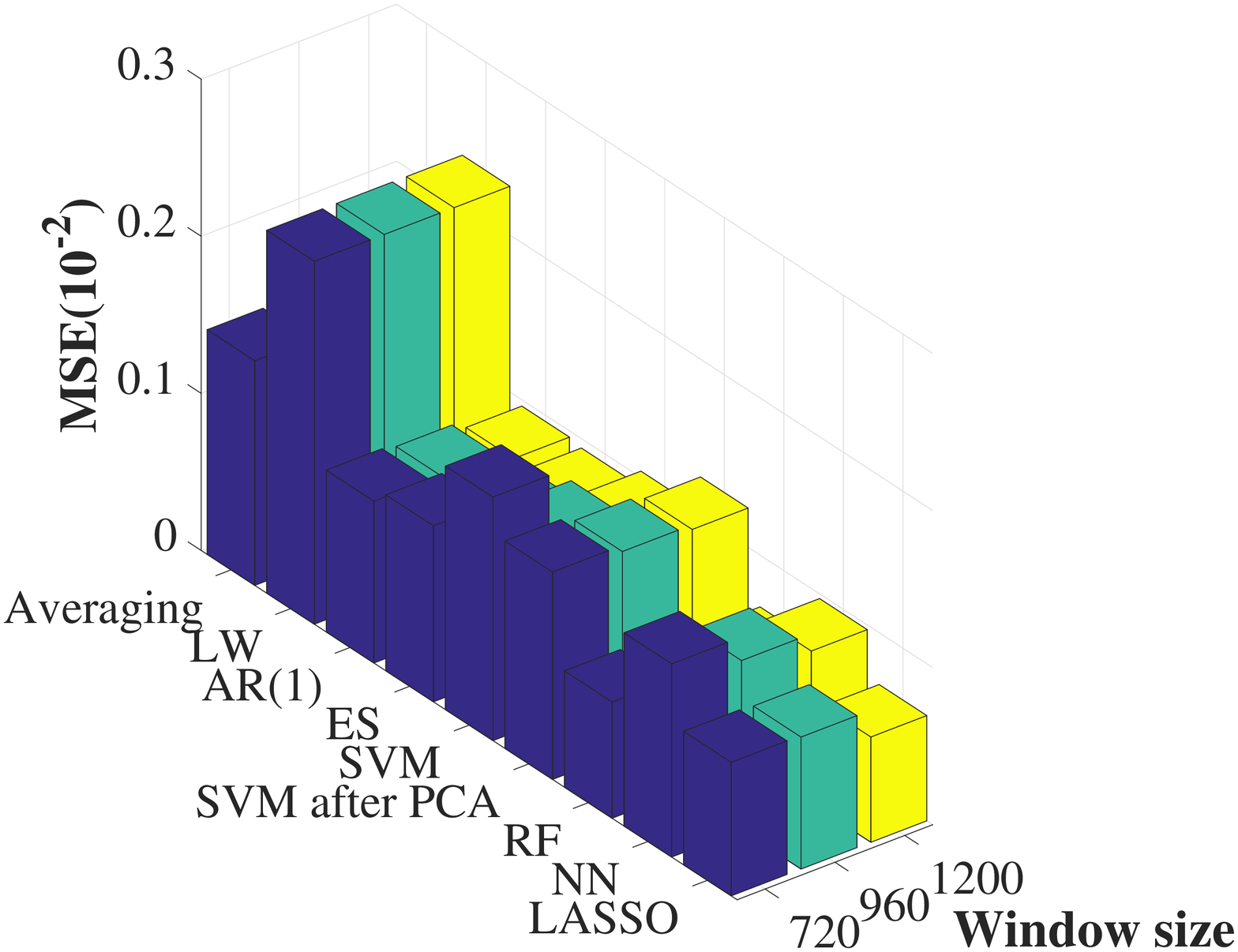}
		\subcaption{Average MSE.}
		\vspace{4ex}
	\end{minipage}
	\begin{minipage}[b]{0.4\linewidth}
		\centering
		\includegraphics[
		width=1\textwidth]{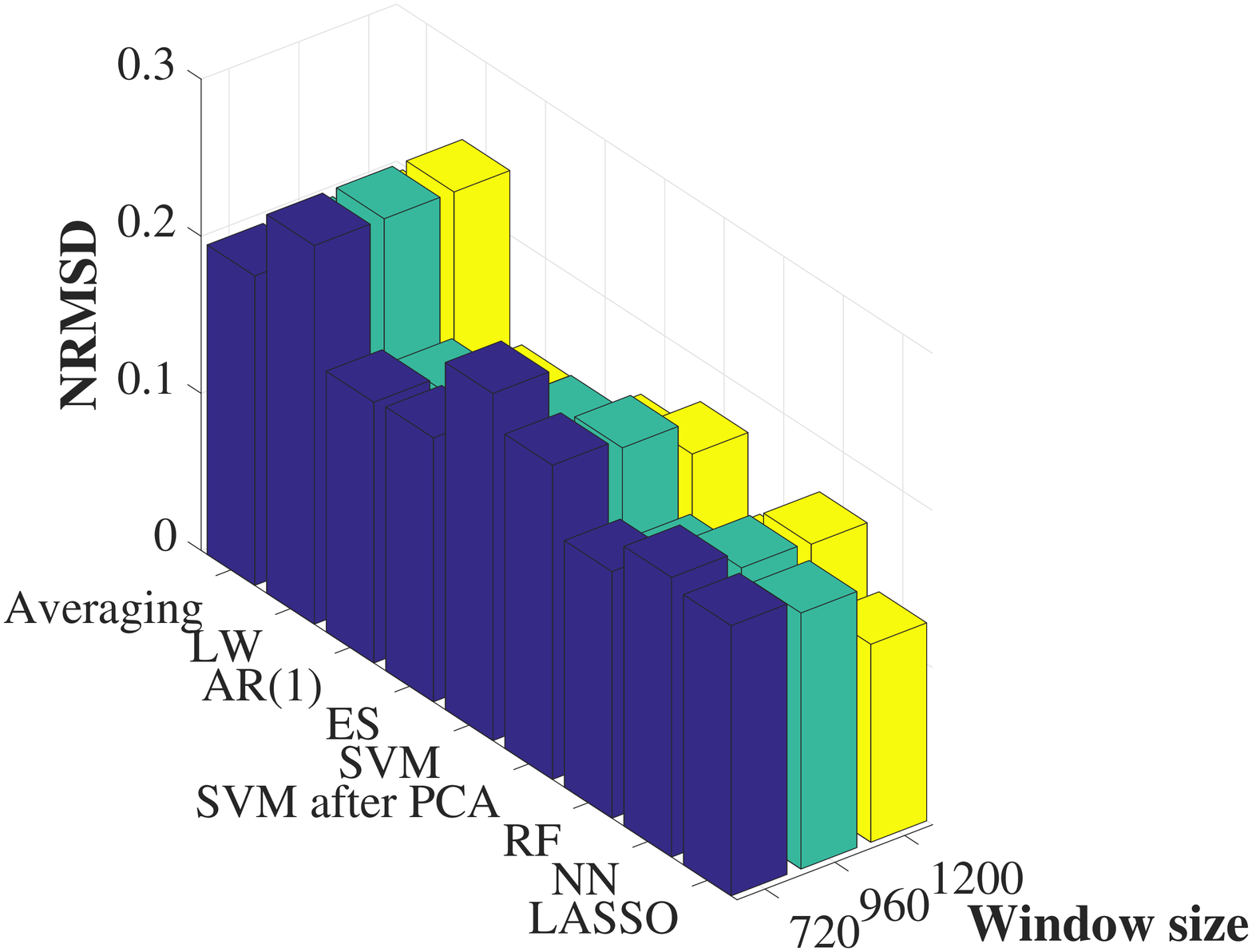}
		\subcaption{Average NRMSD.}
		\vspace{4ex}
	\end{minipage}
	\caption{Results from different error metrics for various models.}
	\label{visComp}
\end{figure*}

We visualize the comparison in terms of MAPE in Fig. \ref{compare}, with 50 users in the training set. In the simulations, the forecast horizon is one hour and the shown errors are the averaged medians over the users. Detailed comparison results based on MAPE, MAE, MSE and NRMSD are presented in Table \ref{MAPEall}, Table \ref{MAEall}, Table \ref{MSEall} and Table \ref{NRMSDall}. In Table \ref{MAPEall} to Table \ref{NRMSDall}, the subscript $W$ denotes the window size. Result from linear regression after PCA is outside the range of the results from the other prediction methods shown in Fig. \ref{compare} and is not presented. These results are visualized in Fig \ref{visComp}, where a smaller error indicates a better predictive performance. In addition, there is a constraint for the minimum number of samples in order to have a well behaved LASSO estimator. This lower bound is of order O($s$log$p$)~\cite{lassomartin}, where $s$ is the cardinality of the true support and $p$ is the total number of regressors. However, since we do not know the level of sparsity $s$ in advance, the number of samples should be at least O($p$log$p$), which is around 600. Nevertheless, since the columns in the regressor matrix are correlated, we resort to a training size larger than the theoretical bound. In the mean time, we cannot include too many samples into the training set to avoid losing stationarity. Here unit root test is applied to test stationarity. To these two ends, we experiment the training sizes of 720, 960 and 1200 samples and compare the respective performances by the proposed prediction methods.

\begin{table}[!ht]
	\renewcommand{\arraystretch}{1.3}
	\caption{Average MAPE cross the 50 users from the dataset for the prediction methods. Subscript $W$ denotes the window size.}
	\label{MAPEall}
	\centering
	\begin{tabular}{|c|c|c|c|}
		\hline
		\bfseries Method &\bfseries $\text{MAPE}_{W = 720}$ &\bfseries $\text{MAPE}_{W = 960}$ &\bfseries $\text{MAPE}_{W = 1200}$  \\
		\hline
		Averaging & 0.258 & 0.258 & 0.258\\
		\hline
		LW & 0.346 & 0.346 & 0.346\\
		\hline
		AR(1) & 0.249 & 0.245 & 0.237\\
		\hline
		ES & 0.250 & 0.244 & 0.236\\
		\hline
		SVM & 0.327 & 0.282 & 0.264\\
		\hline
		SVM after PCA & 0.305 & 0.304 & 0.285\\
		\hline
		RF & 0.211 & 0.204 & 0.203\\
		\hline
		NN & 0.287 & 0.267 & 0.256\\
		\hline
		LASSO & 0.231 & 0.229 & 0.206\\
		\hline
	\end{tabular}
\end{table}


\begin{table}[!ht]
	\renewcommand{\arraystretch}{1.3}
	\caption{Average MAE cross the 50 users from the dataset for the prediction methods. Subscript $W$ denotes the window size.}
	\label{MAEall}
	\centering
	\begin{tabular}{|c|c|c|c|}
		\hline
		\bfseries Method &\bfseries $\text{MAE}_{W = 720}$ &\bfseries $\text{MAE}_{W = 960}$ &\bfseries $\text{MAE}_{W = 1200}$  \\
		\hline
		Averaging & 0.0373 & 0.0373 & 0.0373\\
		\hline
		LW & 0.0540 & 0.0540 & 0.0540\\
		\hline
		AR(1) & 0.0322 & 0.0318 & 0.0311\\
		\hline
		ES & 0.0340 & 0.0334 & 0.0322\\
		\hline
		SVM & 0.0393 & 0.0338 & 0.0322\\
		\hline
		SVM after PCA & 0.0363 & 0.0357 & 0.0368\\
		\hline
		RF & 0.0273 & 0.0265 & 0.0260\\
		\hline
		NN & 0.0360 & 0.0336 & 0.0311\\
		\hline
		LASSO & 0.0291 & 0.0290 & 0.0260\\
		\hline
	\end{tabular}
\end{table}


\begin{table}[!ht]
	\renewcommand{\arraystretch}{1.3}
	\caption{Average MSE cross the 50 users from the dataset for the prediction methods. Subscript $W$ denotes the window size.}
	\label{MSEall}
	\centering
	\begin{tabular}{|c|c|c|c|}
		\hline
		\bfseries Method &\bfseries $\text{MSE}_{W = 720}$ &\bfseries $\text{MSE}_{W = 960}$ &\bfseries $\text{MSE}_{W = 1200}$  \\
		\hline
		Averaging & 0.00143 & 0.00143 & 0.00143\\
		\hline
		LW & 0.00231 & 0.00231 & 0.00231\\
		\hline
		AR(1) & 0.00103 & 0.00101 & 0.00096\\
		\hline
		ES & 0.00113 & 0.00105 & 0.00094\\
		\hline
		SVM & 0.00155 & 0.00115 & 0.00104\\
		\hline
		SVM after PCA & 0.00132 & 0.00128 & 0.00125\\
		\hline
		RF & 0.00074 & 0.00070 & 0.00067\\
		\hline
		NN & 0.00123 & 0.00108 & 0.00097\\
		\hline
		LASSO & 0.00085 & 0.00084 & 0.00067\\
		\hline
	\end{tabular}
\end{table}

\begin{table}[!ht]
	\renewcommand{\arraystretch}{1.3}
	\caption{Average NRMSD cross the 50 users from the dataset for the prediction methods. Subscript $W$ denotes the window size.}
	\label{NRMSDall}
	\centering
	\begin{tabular}{|c|c|c|c|}
		\hline
		\bfseries Method &\bfseries $\text{NRMSD}_{W = 720}$ &\bfseries $\text{NRMSD}_{W = 960}$ &\bfseries $\text{NRMSD}_{W = 1200}$  \\
		\hline
		Averaging & 0.197 & 0.197 & 0.197\\
		\hline
		LW & 0.241 & 0.241 & 0.241\\
		\hline
		AR(1) & 0.166 & 0.156 & 0.135\\
		\hline
		ES & 0.168 & 0.153 & 0.125\\
		\hline
		SVM & 0.221 & 0.182 & 0.153\\
		\hline
		SVM after PCA & 0.200 & 0.194 & 0.173\\
		\hline
		RF & 0.157 & 0.143 & 0.126\\
		\hline
		NN & 0.178 & 0.167 & 0.165\\
		\hline
		LASSO & 0.172 & 0.163 & 0.126\\
		\hline
	\end{tabular}
\end{table}

From Fig. \ref{compare}, we see that RF yields the smallest prediction error in terms of MAPE. This is not unexpected since RF is a complex method that is highly nonlinear. Surprisingly, our proposed LASSO method matches the performance of RF when window sizes are increased. With a sliding window with length of 720 hours, RF has an average MAPE of 0.211 and LASSO has an average MAPE of 0.231. When we increase the size of the sliding window to include 1200 hours' consumption data, RF has an average MAPE of 0.203 and LASSO has an average MAPE as 0.206. At this stage, operators may prefer to use LASSO since it is as accurate as RF but has much lower model complexity. For example, using LASSO operators can understand the most significant hours in the past that influence the prediction. In addition, the coefficients in LASSO can be understood as sensitivities of future prediction on past consumption, due to the linear structure of LASSO. In contrast, the coefficients in RF do not have such ``causal'' structure. In addition, LASSO can be computed much faster than RF.

Besides these two prediction methods, SVM with either the original features or the reduced features are average in their prediction performances. NN does not outperform other methods. We believe this is due to the lack of data. Since we only have time series consumption data without any other side information, only a shallow NN (a single hidden layer) is trained to avoid overfitting and other computational issues.

The nonparametric methods, LW and simple averaging method, return higher MAPEs compared to machine learning techniques. However, based on our dataset, simple averaging using ten previous day's data still outperforms LW (using data from last week only) because the noises are leveraged by adding more consumption data into account. Since these nonparametric methods do not depend on the size of the sliding window, the MAPEs are the same across different sliding window sizes. What is more, these nonparametric methods are powerful when the training set is small (a smaller sliding window size). However, when more data are available to be included into the training set, the performances of these methods do not get improved.

In addition, the time series analysis methods, namely AR(1) and ES, yield moderate MAPE. However, it is still not as good as either RF or LASSO. In conclusion, LASSO achieves as good as the performance from RF, and other methods return inferior prediction performances. Details are shown in Table \ref{MAPEall}. Table \ref{MAEall}, Table \ref{MSEall} to Table \ref{NRMSDall} summarize more results of the comparison based on different metrics. In all, LASSO is shown to be consistently the among the best methods and can provide performances that are as good as advanced machine learning techniques.

\begin{figure}[!t]
	\centering
	\includegraphics[width=1\columnwidth]{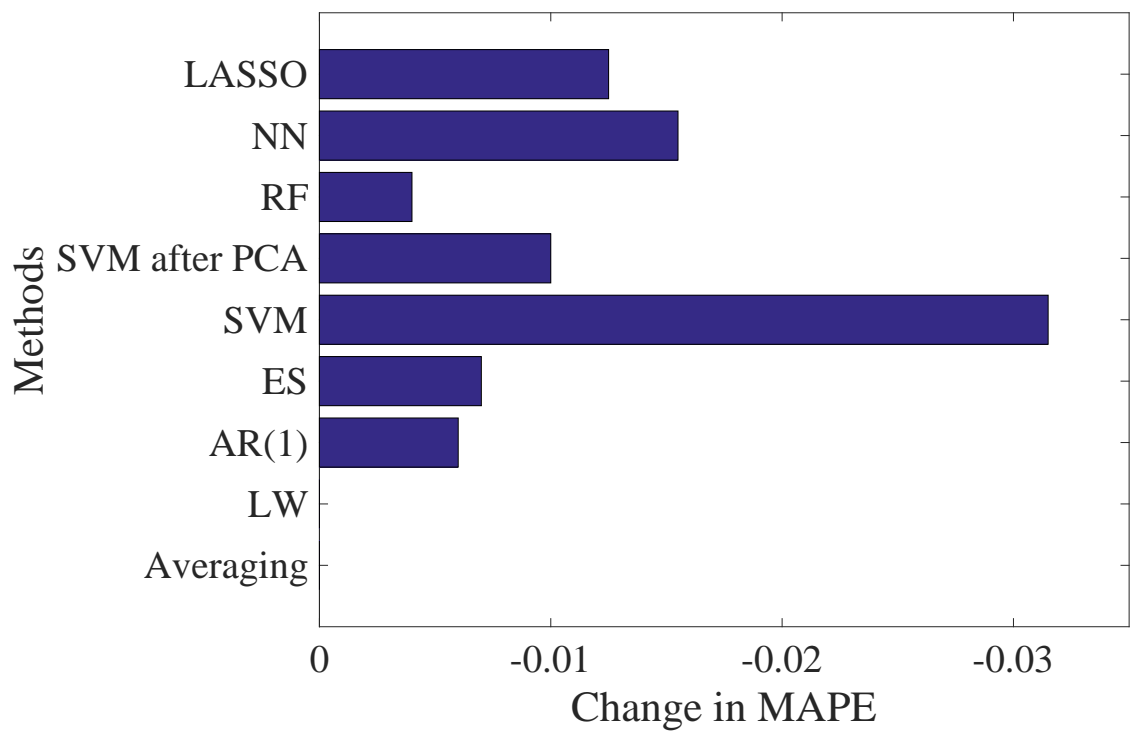}
	\caption{Average change of MAPE with an increase of window size.}
	\label{Sensitivity}
\end{figure}

Besides the above observations on predictive performance of various models, we can also see a difference in model sensitivity with respect to an increase in window size (a larger training set). The results are shown in Fig \ref{Sensitivity}, in terms of MAPE. 

As can be seen from Fig \ref{Sensitivity}, LW and averaging have no sensitivity to window size. This is because they are both an averaged sum of a fixed history and are not parametric. AR(1) and ES have similar sensitivities since they are both autoregressive models. SVM originally has a high sensitivity to window size because we include many features into the model and fix a linear kernel. However, this may not capture correctly the underlying consumption behavior. However, with dimension reduction from PCA, the sensitivity from SVM is improved. Similarly, NN and LASSO do not have a negligible sensitivity towards window size because of the large number of features in the model but since they are more powerful models than SVM, the sensitivity is not as high as the latter. In addition, RF falls into the scheme of ensemble learning algorithm and its variance should be smaller than the other parametric models. Therefore its sensitivity towards window size is small.

We then compare the averaging computational time for the aforementioned methods. The simulations are tested on a MacBook with 2.7 GHz Intel Core i5 processor and 8GB 1867 MHz DDR3 memory. The empirical computational time for each prediction method is shown in Table \ref{timeAll}.

As can be seen from Table \ref{timeAll}, Simple averaging, LW, AR(1) and ES are most time efficient but at the sacrifice of prediction accuracy.  Linear regression with PCA is moderately efficient but provides the worst prediction accuracy. RF yields the worst computational time because it needs to be trained on the repeated bootstrapping samples. NN is also not preferable because for a single layer model it already takes up to more than 14 seconds to train over a single sliding window.
LASSO beats both SVM with either the original regressor matrix or the principle components in terms of time efficiency. Combining the observations from Table \ref{timeAll} with Fig. \ref{compare}, we can see that for univariate time series analysis, LASSO best trades off between prediction accuracy and time efficiency.

Based on the observations from Fig. \ref{compare} and Table \ref{timeAll}, the complexity grows as the prediction model varies from simple averaging method up to RF and consequently the computational time grows accordingly. It is even more time consuming when we try to the tune the hyper-parameters in RF, for example, the number of features to bootstrap at each split of the tree, total number of trees, etc...What is more, although RF provides the most accuracy, it is the hardest to be understood in terms of the estimated parameters from the model, i.e., the decision boundary at each split of the tree node for each tree. Since throughout this paper the focus is on one hour ahead prediction, this complexity ruins the performance of the model and is not preferred. In contrast, LASSO yields a competing prediction performance with RF and has only one hyper-parameter ($\lambda$) to tune. The optimal value of this hyper-parameter can be determined from cross-validation and is easy to implement. Thus it is the most preferable method for univariate analysis based on our dataset.

Table \ref{features} further consolidates this preference in terms of complexity of input data. Feature dimension is defined as the number of input variables in the model. Simple averaging method has a dimension of 10 because it is a average of consumption data during the same hour for the past ten days. LW only needs one data point during last week to make prediction. AR(1) and ES also have a much smaller input space but their prediction performances are not as good as either LASSO or RF. For methods associated with dimension reduction by PCA, the feature dimension is the number of the reduced features for the model, thus around 20 in our case. We restrict the feature dimension as 24 for NN since a higher feature dimension will require more iterations in the algorithm to reach convergence. Other than that, for SVM, RF and LASSO, the input data has a dimension of 240 which stands for the total number of hours in the past ten days. With the same input complexity, LASSO yields the most efficient computational time from Table \ref{timeAll}.

Based on the discussions from Fig. \ref{compare}, Table \ref{timeAll} and Table \ref{features}, LASSO is the best choice in terms of model efficiency and model simplicity.

\begin{table}[!ht]
	\renewcommand{\arraystretch}{1.3}
	\caption{Average computational time per user per sliding training window for each prediction method.}
	\label{timeAll}
	\centering
	\begin{tabular}{|c|c|}
		\hline
		\bfseries Prediction method &\bfseries Time (in seconds)  \\
		\hline
		Averaging & less than 0.0001\\
		\hline
		LW & less than 0.0001\\
		\hline
		AR(1) &  0.005 \\
		\hline
		ES & 0. 093\\
		\hline
		PCA + linear regression & 0.175 \\
		\hline
		SVM & 9.347 \\
		\hline
		SVM + PCA & 9.522 \\
		\hline
		RF & 21.562\\
		\hline
		NN & 14.416\\
		\hline
		LASSO & 0.220 \\
		\hline
	\end{tabular}
\end{table}

\begin{table}[!ht]
	\renewcommand{\arraystretch}{1.3}
	\caption{Feature dimension for different prediction methods.}
	\label{features}
	\centering
	\begin{tabular}{|c|c|}
		\hline
		\bfseries Prediction method &\bfseries Feature dimension  \\
		\hline
		Averaging & 10\\
		\hline
		LW & 1\\
		\hline
		AR(1) &  1 \\
		\hline
		ES & at most 3\\
		\hline
		PCA + linear regression & around 20 \\
		\hline
		SVM & 240 \\
		\hline
		SVM + PCA & around 20\\
		\hline
		RF & 240 (split the tree 500 times.)\\
		\hline
		NN & 24 \\
		\hline
		LASSO & 240 (active regressors around 10-20.) \\
		\hline
	\end{tabular}
\end{table}

In addition, the average computational time for computing the optimal pairs using the covariance test is 3.51 seconds on average per each user. We then conclude that it is time efficient to compute the optimal pairs of users. What is more, since the optimal pair assignment is fixed once we compute it, this computational time can be amortized and is nearly negligible.

\subsection{Results for Univariate Time Series Analysis}
Based on the comparison result shown in Section \ref{section_compare}, in this subsection we perform more detailed simulations on the following prediction methods from the methods considered so far:
\begin{itemize}
	\item Simple averaging method, which is a representative method from nonparametric methods;
	\item AR(1), which is a representative method from time series methods;
	\item RF, which is a representative method from machine learning techniques;
	\item LASSO applied in autoregressive model, which is the proposed method in this paper.
	\item SVM with and without PCA, which we use as an illustration of the failure of PCA based on our dataset.
\end{itemize}

\begin{figure}[!t]
	\centering
	\includegraphics[width=1.1\columnwidth]{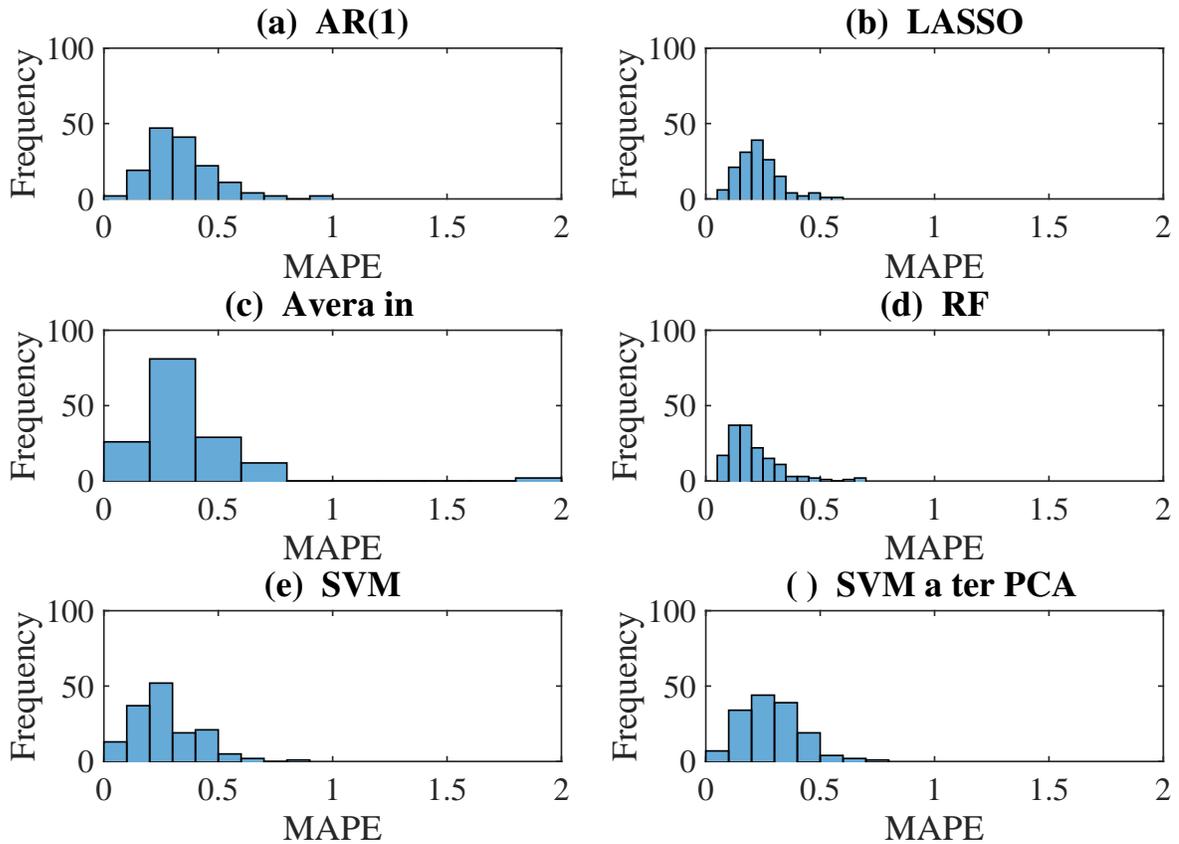}
	\caption{Histogram comparison for univariate time series. Mean of MAPE for AR(1) is 0.339, for LASSO 0.225, for simple averaging method 0.359, for RF 0.195 and for SVM with and without PCA is 0.264 and 0.291 respectively. Standard deviation of MAPE for AR(1) is 0.151, for LASSO 0.094, for simple averaging method 0.237, for RF 0.109, for SVM 0.142 and for SVM after PCA is 0.129. }
	\label{unicompare}
\end{figure}

The simulation results associated with the covariance test are going to be introduced later in Section \ref{section_multi}. The results of the prediction methods besides covariance test are shown in Fig. \ref{unicompare} and Table \ref{table:2}, for a random selection of 150 users from the dataset. We use MAPE as the criterion to compare prediction performances because it is a normalized metric. Again, the MAPE shown is the median over all the sliding windows and the forecast horizon is one. As can be seen in Fig. \ref{unicompare}, LASSO type regression reduces the variance of MAPE for these 150 users in the dataset. Its mean is also reduced. A comparison of the aforementioned methods is summarized in Table \ref{table:2}.

From Fig. \ref{unicompare}, we see that MAPE varies on a user to user basis. For example, from Fig. 2(a), using AR(1) model to predict consumption will yield a MAPE greater than 50\% for some of users. What is more, there is one extreme case for simple averaging method shown in Fig. 2(c) that MAPE reaches 200\%. This extreme value of MAPE suggests a rather unfavarable prediction result, even worse than simply predicting zero (which has a MAPE as 100\%.). However, for both methods, a MAPE close to zero can still be achieved for other users. Besides these methods, SVM achieves a moderate deviation of MAPE but it is still higher that by LASSO and RF. Comparing Fig. 2(a), Fig. 2(c) with Fig. 2(b), Fig. 2(d), Fig. (e) and Fig. (f), we see that LASSO and RF have MAPEs that are less spread between 0\% and 100\%. In addition, LASSO and RF have a smaller mean of MAPE than that from the other prediction methods. More detailed statistics are shown in Table \ref{table:2}. What is more, comparing Fig. 2(e) and Fig. 2(f), PCA does not improve the performance of prediction models such as SVM, as we have discussed in Section V. B. Here, prediction by SVM using the original features (historical consumption data) as inputs achieves a smaller MAPE than that after dimension reduction by PCA. SVM after PCA has a MAPE of 0.291 against a MAPE of 0.264 by SVM alone.

In addition, besides the improved prediction performance compared to AR(1) and simple averaging method, LASSO also provides some straightforward understandings of the model. Particularly, we include as much as 240 lag orders for LASSO, which includes all the historical data for the previous ten days. Taking user No.1 as an example, LASSO selects 16 non zero lag orders, according to 10-fold CV with a sequence of decreasing $\{\lambda_k\}$. The lag orders that LASSO picks are 1, 2, 5, 6, 16, 22, 23, 24, 48, 143, 144, 160, 191, 216, 238, 240. This pattern reflects that LASSO not only selects the most recent lag orders (which is similar to AR(1)), but the lag orders roughly at interval of one day, i.e., 24 hours as well (which is similar to simple averaging). The coefficients for the lag orders also have different scales. In the example of user No.1, the most recent lag orders given the largest coefficient ($\beta_1$ = 0.259). The second largest coefficient is given to lag order 24 ($\beta_{24}$ = 0.187). The rest of the coefficients are scaled between 0.01 to 0.06.

Note that LASSO will return selected lag orders adaptively for different users. For user No.10, the lag orders that LASSO picks up for this user are 1, 23, 24, 120, 138, 144, 168, 239, 240. In this case, the three most recent lag orders are given the largest coefficients, where $\beta_1$ = 0.579, $\beta_{23}$ = 0.123 and $\beta_{24}$ = 0.103. The rest of the coefficients scale at around 0.02. So for this particular user, LASSO picks up less historical data at the same hour than user No. 1, at roughly one day ago, five days ago, six days ago, seven days ago and ten days ago. This is different from simple averaging, since LASSO does not pick up the lag order 48 or 72, which are more recent lags than 120. Still, however different lag orders that LASSO picks for each user, we can observe a clear pattern of multiples of 24 hours, indicating user's consumption behavior.

In all, the proposed model based on LASSO is useful in practice because of its simplicity and linear nature. Given the set of coefficients, utilities and planners can easily identify and attribute the impact of different features on load consumption. For example, suppose for a given user, our method concludes that its load at time $t$ is mostly determined by the past load at times $t-1$, $t-3$ and $t-24$. This shows that this user has some short term behavior in the one hour range, medium term behavior in the 3 hour range, and a daily cycle partner that repeats every 24 hours. This information can be interpreted by utilities to better understand this user's consumption patterns and potentially identify appropriate demand programs.

\subsection{Results for Pairing the Users}\label{section_multi}

As described in Section \ref{multi}, we include historical consumption data from other users to fit the current user's data. We compare the results of the two methods discussed as follows:
\begin{enumerate}
	\item Reference method: LASSO selection in autoregressive model using one user's own data.
	\item Pairing by covariance test: the optimal pairing for a user is selected by the best user fitted to the residuals obtained from this user' consumption data after LASSO in univariate autoregressive model. This best user corresponds to the first regressor in $\bm{\xi}$ entering the active set when its associated $p$-value is small. The lag orders for this user are selected by LASSO in the reference method, but the lag order for its paired user is fixed to one.
\end{enumerate}

\begin{figure}[!t]
	\centering
	\includegraphics[width=0.9\columnwidth]{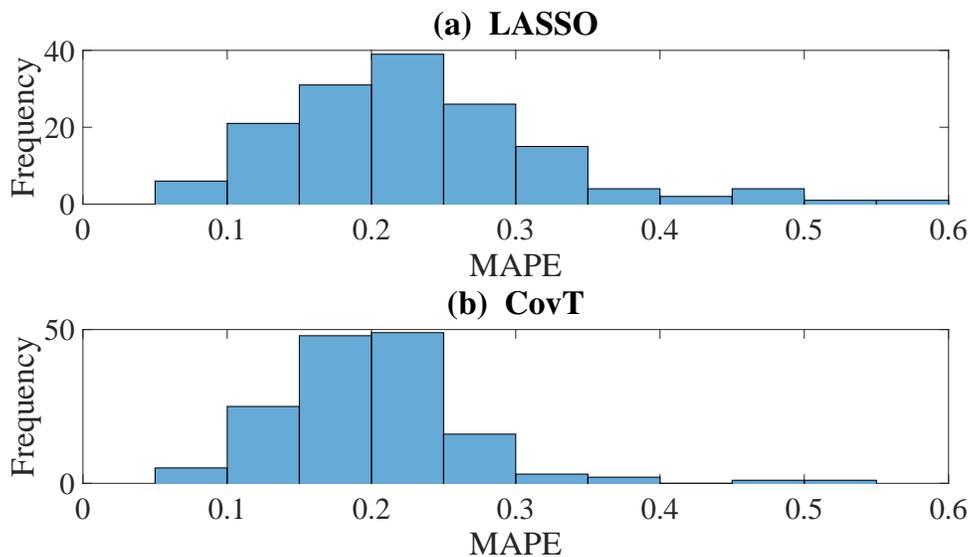}
	\caption{Histogram comparison for multivariate time series, where covT stands for covariance test. Mean MAPE for LASSO is 0.225 and for performing covariance test after LASSO is 0.209. Standard deviation of MAPE has also reduced by 7\% from LASSO to performing covariance test after LASSO. }
	\label{multicompare}
\end{figure}

The results for these two methods are in Fig. \ref{multicompare} and Table \ref{table:2}, with a forecast horizon as one. The results are based on the same 150 users chosen from the dataset. It can be seen from Fig. \ref{multicompare} that with the covariance test that includes another useful user into the model, the prediction performance is enhanced. The mean of the MAPE on the test set is improved by 7\% and the variance by 6\%. Note that the selection of another user into the model is not mutual, for example, for user No.14, user No.137 is selected by covariance test whereas for user No.137, user No.50 is selected. This is essentially different from bivariate autoregressive model where pairs of users are included into the model simultaneously. Here, the pairing of the users is adaptive according to the covariance statistic defined in (\ref{T}).


\begin{table}[!ht]
	\renewcommand{\arraystretch}{1.3}
	\caption{Results for the prediction methods, where CovT stands for covariance test and Ave stands for simple averaging method.}
	\label{table:2}
	\centering
	\begin{tabular}{|c|c|c|c|c|c|c|c|}
		\hline
		\bfseries  &\bfseries Ave &\bfseries  AR(1)& \bfseries LASSO & \bfseries RF & \bfseries SVM & \bfseries CovT\\
		\hline
		mean (MAPE) & 0.359& 0.339 & 0.225 & 0.195 & 0.264 & 0.209\\
		\hline
		sd (MAPE) & 0.237& 0.151& 0.094 & 0.109 & 0.142 & 0.088\\
		\hline
	\end{tabular}
\end{table}

As can be seen from Table \ref{table:2}, the proposed methods greatly improve the prediction performance by reducing the relative prediction error, thus will provide more reliable results on consumption prediction. The proposed prediction methods in this paper can help compare the user behavior with and without demand response incentives. In the future, we will use the results in this paper to investigate the demand response effect of users by comparing their actual consumption data and predicted consumption data.

\section{Conclusions and Future Works}\label{conclusion}
In this paper, we propose a sparse autoregressive model to predict the electricity consumption of individual users. First, we adopt LASSO to recover the sparsity in linear regression models. LASSO selects the most recent lag orders and important lag orders with close multiples of 24 hours, which reveals user consumption patterns. Second, we improve prediction accuracy of a particular user by leveraging other user's historical data, where we use the covariance statistic to test if inclusion of another user is significant to explain the fitted residual from univariate LASSO-type regression. Third, we give a comprehensive analysis on the proposed methods against several available predictive models in literature. We observe that LASSO best trades off between model complexity and prediction performance. It has the least predictive error among the linear models and yields a competitive performance as compared to ensemble methods such as random forests. Further simulation results show that LASSO with covariance test outperforms both simple averaging method and auto regressive method with order 1 for individual consumption prediction, improving the prediction performance by 38.4\% in terms of MAPE.

\bibliographystyle{IEEEtran}	
\bibliography{IEEEabrv,paperref}

\appendices
\section{Statistics concepts}

Since many statistical terms appear throughout the paper, we therefore give more explanations to them in this appendix.

\subsection{Univariate autoregressive model}
An \emph{univariate autoregressive model of consumption data $y_t$ is expressed in the following form:}
\begin{equation}
y_{t} = \beta_{0}+ \sum_{i=1}^{I}{\beta_{i}y_{t-i}} + \epsilon_{t}.
\end{equation}
where $y_{t}$ is a scalar at each time $t$. It simply refers to the autoregressive models for one dimensional time series data.

\subsection{Vector autoregressive model}
A \emph{vector autoregressive model} aims to jointly predict multiple users' future demand by their past consumption data. It is expressed in the following form:
\begin{equation}
\bm{z}_{t} = \bm{A}_{0}+ \sum_{i=1}^{I}{\bm{A}_{i}\bm{z}_{t-i}} + \bm{\epsilon}_{t}.
\end{equation}
where $\bm{z}_t$ is a consumption vector at each time $t$ for multiple users. $A_{i}$ is the parameter matrix at each lag order $i$ associated with these users' past consumption data. Therefore, vector autoregressive model refers to the autoregressive model for multi-dimensional time series data.

\subsection{Out-of-sample error}
\emph{Out-of-sample error} depicts prediction error of the statistical model. More specifically, the out-of-sample error is the deviation from consumption prediction to the real consumption that is not used to train the model by the operator. For example, in OLS estimation, the objective is to minimize the sum of squared error for the samples in the training set. {Out-of-sample error} is thus the estimation error in the test set.

\subsection{AIC/BIC}
\emph{AIC} (Akaike information criterion) is a measure of relative performance of statistical models. It is similar to traditional error metrics such as MSE only that model complexity is involved in its computation. AIC computed as:
\begin{equation}
\text{AIC} = 2k - \ln L
\end{equation}
where $k$ is the number of free parameters in the model and $L$ is the likelihood.

Similarly, \emph{BIC} (Bayesian Information Criterion) is also a measure of relative performance of statistical models. It is defined as the following:
\begin{equation}
\text{BIC} = -2 \ln L + k \ln n
\end{equation}
where $n$ is the number of data points in the dataset.

\subsection{Active set}
An \emph{active set} is the set of all non zero coefficients estimated at a fixed value of the tuning parameter $\lambda$ in LASSO. In this paper, the active set for one particular user includes the indices of users that are potentially useful in predicting this user's future demand.

\subsection{Tailstop criterion}
\emph{Tailstop criterion} is introduced in~\cite{sequentialsig} and is used as a stopping criterion in a sequence of significance tests. For more details, please refer to~\cite{sequentialsig}.

\subsection{Covariance statistic}
\emph{Covariance statistic} is used in the significance test for LASSO. More specifically, it is the statistic that tests whether the inclusion of the variables are significant or not. In this paper, it is used to select the index of the user whose past consumption data is most useful in predicting one particular user's future demand. The larger the statistic, the more significant the selected user's influence is on the prediction.

\section{Model comparison using AIC/BIC}
Besides the error metrics introduced in Section \ref{simulation}. A, AIC and BIC are two other criteria to model selection. We therefore use AIC and BIC to compare the autoregressive models with order up to 5 and the proposed sparse model obtained by LASSO. Since the objective for both models is to minimize the RSS (Residual Sum of Squares), AIC/BIC can be reduced to:
\begin{equation}
\text{AIC} = 2k + n\ln(RSS) + C
\end{equation}
where C is a constant across models, and
\begin{equation}
\text{BIC} = n \ln (RSS/n) + k\ln n
\end{equation}

For autoregressive models, $k$ is equal to the number of lag orders in the model. For LASSO, $k$ is defined as the level of sparsity of the model.

\begin{figure}[ht]
	\centering
	\includegraphics[width=0.85\columnwidth]{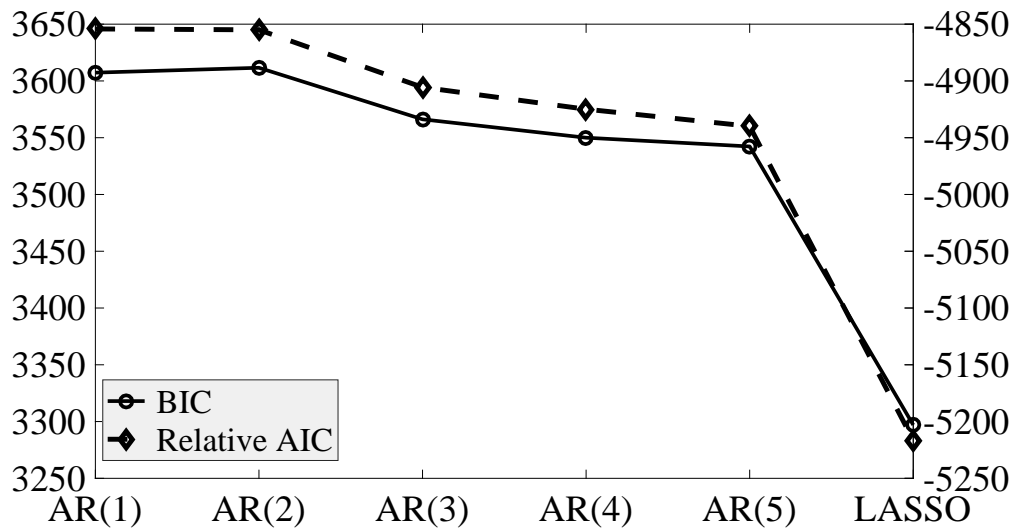}
	\caption{Comparison between different autoregressive models using AIC and BIC.}
	\label{AICBIC}
\end{figure}

A comparison between AR(x) model and LASSO based on AIC/BIC is provided in Table \ref{AICcomp} and in Fig. \ref{AICBIC}, with a randomly selected user from the pool. Note that lower values of AIC and BIC are preferred, however their absolute value does not indicate whether the model is good or not. Therefore AIC and BIC criteria are \emph{relative} measures of the model quality.

\begin{table}[!ht]
	\renewcommand{\arraystretch}{1.3}
	\caption{Comparison between different autoregressive models using AIC and BIC.}
	\label{AICcomp}
	\centering
	\begin{tabular}{|c|c|c|}
		\hline
		\bfseries Method &\bfseries AIC &\bfseries BIC\\
		\hline
		AR(1) & 3645.8+$C$ & -4892.7 \\
		\hline
		AR(2) & 3645.0+$C$ & -4888.4\\
		\hline
		AR(3) & 3594.4+$C$ & -4933.8\\
		\hline
		AR(4) & 3575.1+$C$ & -4950.0 \\
		\hline
		AR(5) & 3560.1+$C$ & -4957.7 \\
		\hline
		LASSO & 3283.5+$C$ & -5203.0 \\
		\hline
	\end{tabular}
\end{table}

As can be seen from Table \ref{AICcomp} and Fig. \ref{AICBIC}, LASSO has both the lowest AIC as well as BIC, which implies that LASSO is the best model among the models presented in Table \ref{AICcomp}. From Table \ref{AICcomp} we observe that the AIC and BIC from AR($x$) models are quite similiar, with only a change of 1.3\%. However, AR(5) has an increase of 23.8\% in out-of-sample error, as shown in Fig. \ref{error}. Therefore AIC/ BIC is not a preferable criteria to compare predictive performance based on our data set.








\end{document}